\documentclass[sigconf]{acmart}
\usepackage{fancyhdr}
\usepackage{lipsum}

\renewcommand\footnotetextcopyrightpermission[1]{}

\AtBeginDocument{%
  \providecommand\BibTeX{{%
    \normalfont B\kern-0.5em{\scshape i\kern-0.25em b}\kern-0.8em\TeX}}}

\def\appendix{\par
\section*{APPENDIX}
\setcounter{section}{0}
 \setcounter{subsection}{0}
 \def\thesection{\Alph{section}} }

\setcopyright{acmcopyright}
\copyrightyear{2018}
\acmYear{2018}
\acmDOI{XXXXXXX.XXXXXXX}

\acmConference[PGAI CIKM 2023]{International Workshop on Personalized Generative AI (@CIKM 2023)}{Oct 22, 2023}{Birmingham, UK}
\acmPrice{15.00}
\acmISBN{978-1-4503-XXXX-X/18/06}

\begin{document}

\title{ChatCounselor: A Large Language Models for Mental Health Support}


\author{June M. Liu}
\authornote{All three authors contributed equally to this research.}
\email{juneliu@connect.hku.hk}
\affiliation{%
  \institution{Hong Kong University}
  \city{Hong Kong SAR}
  \country{China}
}

\author{Donghao Li}
\authornotemark[1]
\email{dlibf@connect.ust.hk}
\author{He Cao}
\authornotemark[1]
\email{hcaoaf@connect.ust.hk}
\affiliation{%
  \institution{Hong Kong University of Science and Technology}
  \city{Hong Kong SAR}
  \country{China}}

\author{Tianhe Ren}
\email{rentianhe@idea.edu.cn}
\affiliation{%
  \institution{The International Digital Economy Academy (IDEA)	}
  \city{Shenzhen}
  \country{China}}

\author{Zeyi Liao}
\email{liao.629@osu.edu}
\affiliation{%
  \institution{Ohio State University}
  \city{Columbus}
  \country{USA}}

\author{Jiamin Wu}
\email{jwubz@connect.ust.hk}
\affiliation{%
  \institution{Hong Kong University of Science and Technology}
  \city{Hong Kong SAR}
  \country{China}}

\renewcommand{\shortauthors}{Liu and Cao, et al.}

\begin{abstract}
This paper presents ChatCounselor, a large language model (LLM) solution designed to provide mental health support. Unlike generic chatbots, ChatCounselor is distinguished by its foundation in real conversations between consulting clients and professional psychologists, enabling it to possess specialized knowledge and counseling skills in the field of psychology. The training dataset, \textit{Psych8k}, was constructed from 260 in-depth interviews, each spanning an hour. To assess the quality of counseling responses, the \textit{counseling Bench} was devised. Leveraging GPT-4 and meticulously crafted prompts based on seven metrics of psychological counseling assessment, the model underwent evaluation using a set of real-world counseling questions. Impressively, ChatCounselor surpasses existing open-source models in the counseling Bench and approaches the performance level of ChatGPT, showcasing the remarkable enhancement in model capability attained through high-quality domain-specific data.

\end{abstract}

\maketitle

\section{Introduction}
In modern society, mental health has become an increasingly critical concern \cite{prince2007no}. As we navigate through a fast-paced and interconnected world, individuals face numerous challenges that can impact their psychological well-being. 
To ensure mental well-being, timely early psychological intervention and diagnosis are crucial. However, professional psychological counselors often face challenges in providing immediate assistance when mental health issues arise due to high costs, complicated appointment procedures, and limited resources.

The emergence of large language models (LLMs) offers a potential solution by utilizing LLMs to make initial assessments and provide early interventions to individuals. LLMs can leverage their vast knowledge and language processing capabilities to engage in conversations and offer support to those in need. However, existing LLMs solutions such as GPT4 \cite{gpt4} and PaLM \cite{palm} primarily focus on general domains and are trained on widely available generic datasets. These models often lack specific training and expertise in the field of psychological counseling.

This paper introduces  ChatCounselor—--a customized LLaMA-7B model fine-tuned with counseling domain instruction data, named Psych8k. 
Unlike current psychological counseling datasets \cite{psyqa}, which are obtained by scraping online forums, where participants lack professional backgrounds in psychological counseling, Psych8k is exclusively generated by licensed psychological counselors, guaranteeing its high quality.
Specifically, we collected counseling dialogue from 260 individuals and employed GPT-4\cite{gpt4} as an extractor and filter to construct 8,187 instruction paris. 

In addition, to evaluate its counseling ability, we build counseling
Bench, which consists of a comprehensive set of metrics, specifically tailored to the LLM counseling domain, 229 real-world questions for evaluation, and a GPT4-powered automatic evaluation pipeline.
In this benchmark, ChatCounselor shows improved performance against its base model, showing a small amount of high-quality data could improve performance in a specific domain. 

\begin{figure}[htbp]
  \centering
  \includegraphics[scale=0.25]{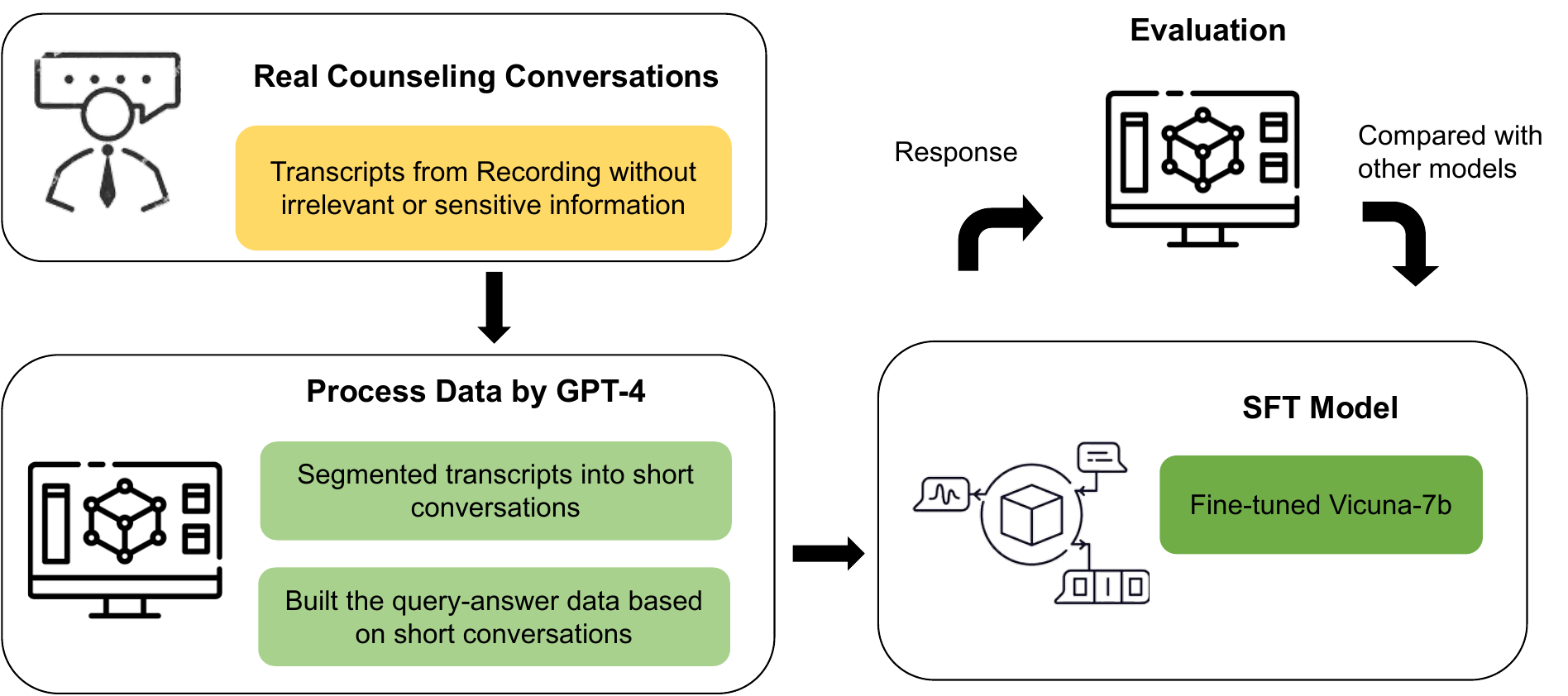}
  \caption{Pipeline of ChatCounselor}
  \label{fig:pipeline}
\end{figure}

Figure\ref{fig:pipeline} provides an overview of our paper.

\textit{Our contributions are the following:} 
\begin{itemize}
    \item We have constructed the Psych8k dataset from conversations with 260 different individuals, which can be used to explore personalized psychological counseling assistants.
    \item We propose a benchmark framework that utilizes GPT-4 for automated evaluation, which provides a rapid assessment method for the development of psychological counseling assistants.
    \item We found that the model fine-tuned with Psych8k outperforms existing open-source models in the benchmark, demonstrating that a small amount of high-quality data can significantly enhance the performance of psychological counseling assistants.
\end{itemize}

We make our code available on GitHub.\footnote{\url{https://github.com/EmoCareAI/ChatPsychiatrist}}

\section{Related Work}

\subsection{LLMs for psychological service} LLMs find extensive use across multiple domains. Their capacity to generalize from vast text data benefits fields from healthcare\cite{huatuogpt,li2023chatdoctor} to finance\cite{wu2023bloomberggpt}, education\cite{malinka2023educational,susnjak2022chatgpt} to law\cite{cui2023chatlaw,lawgpt}, automating and improving communication, data analysis, and decision-making. 

The utilization of LLMs in counseling and mental health support is an emerging area\cite{singh2023mind}. Researchers have proposed models that facilitated non-professionals to offer effective counseling \cite{fu2023enhancing} or provided peer support \cite{lai2023psy}. Moreover, emotional support conversation (ESC) or dialog systems \cite{liu2021towards} was developed based on Hill’s Helping Skills Theory \cite{hill2009helping} to generate support to reduce daily distress. Further improvement on ESC, including CASE \cite{zhou2022case} that modeled the interaction between cognition and affection for empathetic help, and PAL \cite{cheng2022pal} that considered persona information in emotional support, all facilitated LLMs’ efficacy in helping. Later, SMILE \cite{qiu2023smile} expanded the crowd-sourcing conversations’ data used in the above studies to multi-turn dialogs, making the outputs closer to real-life conversations. The benchmark to measure counseling dialogue’s safety was also developed \cite{qiu2023benchmark}. Based on those findings, our study will focus on condensing the information from real counseling conversations, to make the LLM’s feedback more professional.

\subsection{Personalized large language models}
Personalized large language models (PLLMs) are designed to provide customized response strategies based on individual users. These models learn from user preferences, context, and personal characteristics to generate replies that align with the users' needs and preferences. By employing personalized response strategies, these models can better meet user requirements and provide more targeted and personalized interactive experiences. Recently, researchers have evaluated PLLMs on a diverse range of tasks, including generating news headlines and email subjects \cite{salemi2023lamp}, generating academic prose, and generating ideas \cite{porsdam2023autogen}. In the field of recommendation systems, PLLMs have been proven to possess powerful zero-shot capabilities that surpass existing custom models \cite{sileo2022zero,hou2023large,he2023large}. This paper provides a new question-answering dataset for evaluating PLLMs methods, consisting of 260 individuals with an average of 31 question-answer pairs per individual.

\section{Methodology} 
Our approach centers on refining a LLM that has been trained on a vast general corpus for counseling domain application. We achieve this by integrating authentic counseling data and domain-specific psychological knowledge to augment the conversational quality. To incorporate domain knowledge, we employ instruction tuning combined with autoregressive training. When constructing the dataset, our objective is to ensure that responses are not only informative and professional but also incorporate counseling skills such as reflection, active listening, and interaction. Additionally, we strive to capture the natural tone and demeanor of a counseling professional.

\subsection{Data Collection}
Clinical or therapeutic data were typically collected from real-world conditions or related Q\&A platforms to help the LLM be more professional. For example, ChatDoctor\cite{chatdoctor} employed physician-patient conversation from online medical consultation websites. Similarly, HuatuoGPT\cite{huatuogpt} combined distilled data with real-world Chinese doctor-patient queries and answers to improve the chatbot’s abilities. However, there are some concerns when using such data: 

(1) Some Q\&A datasets\cite{psyqa} crawled from online counseling platforms were not provided by respondents with professional accreditation. (2) Answers from online counseling platforms tended to solve problems as soon as possible, so those answers provided instant suggestions for action. But real-world counseling requires counselor to know more about the client before giving advice; thus it would be better if the chatbot can ask questions and discuss about the client’s personal background and past experiences in the conversation. (3) If collecting data from real-world counseling videos, content should be well-processed since the raw transcripts were usually fragmented. 

To solve the first two concerns, we collected 260 real-life counseling recordings in English, which were all provided by people with accreditation and extensive counseling experience. The raw counseling recordings were transcribed to acquire textual data. Then we processed the texts to make the conversation concise (see 3.2). 

\begin{figure}[htbp]
  \centering
  \includegraphics[scale=0.35]{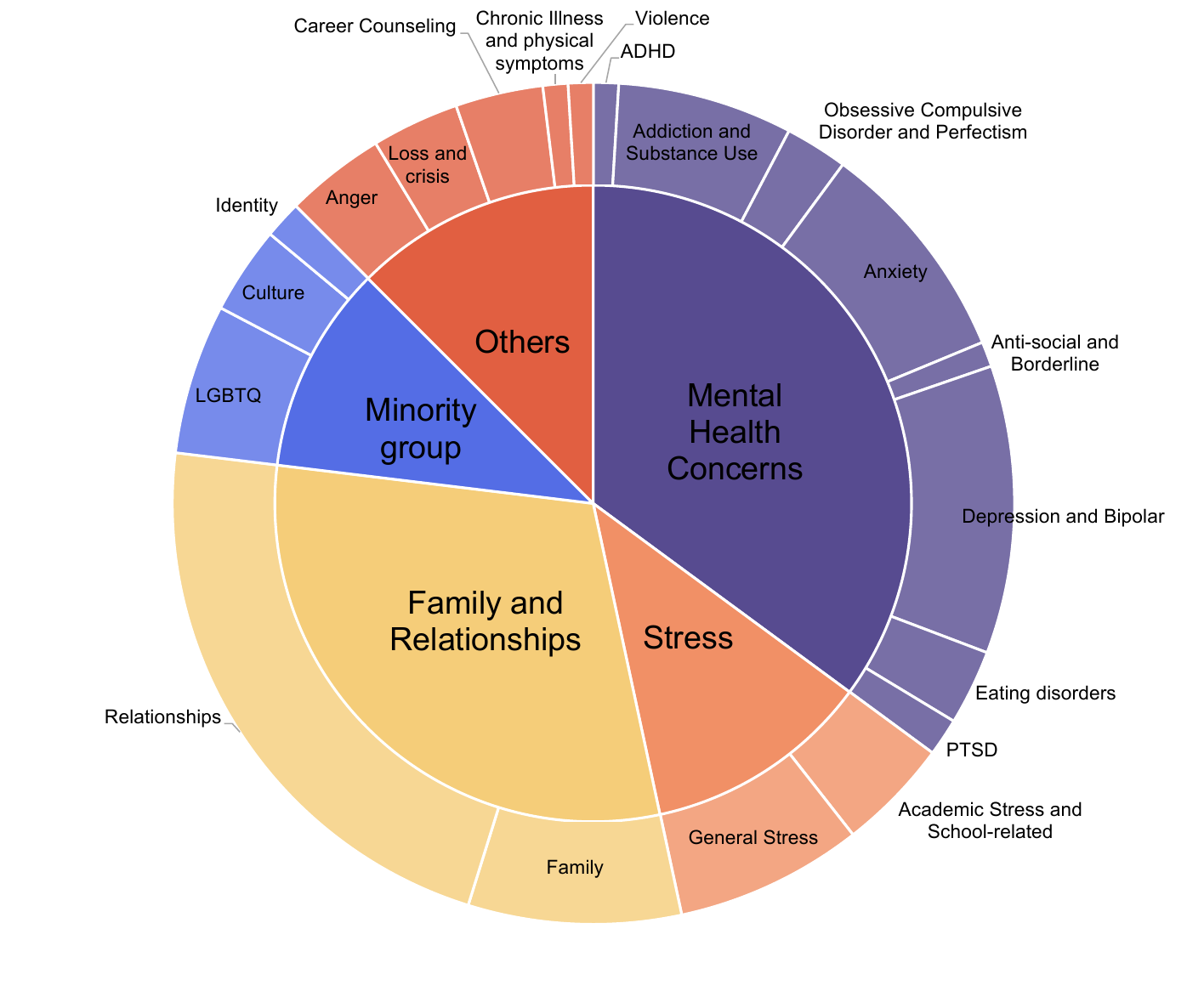}
  \caption{The distribution of mental health-related topics. The inner circle represents 5 major categories, and the outer group represents minor topics.}
  \label{fig:Topic Coverage}
\end{figure}

As for the dataset’s topic coverage, we first included the most prevalent mental health concerns, such as anxiety \cite{thibaut2022anxiety} and depression \cite{mccormick2013stressed}. Then, a large part of the topics was related to stress and relationship since those two were important predictors of mental disorders \cite{slavich2014stress,kansky2018long,li2019intergenerational}. We also considered people from minority groups, such as LGBTQ and different cultural groups. Overall, these conversations cover five major topics. For each major topic, we have further subdivided it into a number of subtopics. (see Fig.\ref{fig:Topic Coverage} for distribution of data). 

\subsection{Data Processing}
\subsubsection{Data Cleaning \& Information Extraction} 

The raw counseling recordings were transcribed to acquire textual data. To maintain privacy and ethical standards, the obtained transcripts were thoroughly cleaned to eliminate any irrelevant or sensitive information. This step involved removing personally identifiable information (PII) and any other content that may disclose the identity of the participants or compromise the integrity of the data. 

\begin{figure}[htbp]
  \centering
  \includegraphics[scale=0.35]{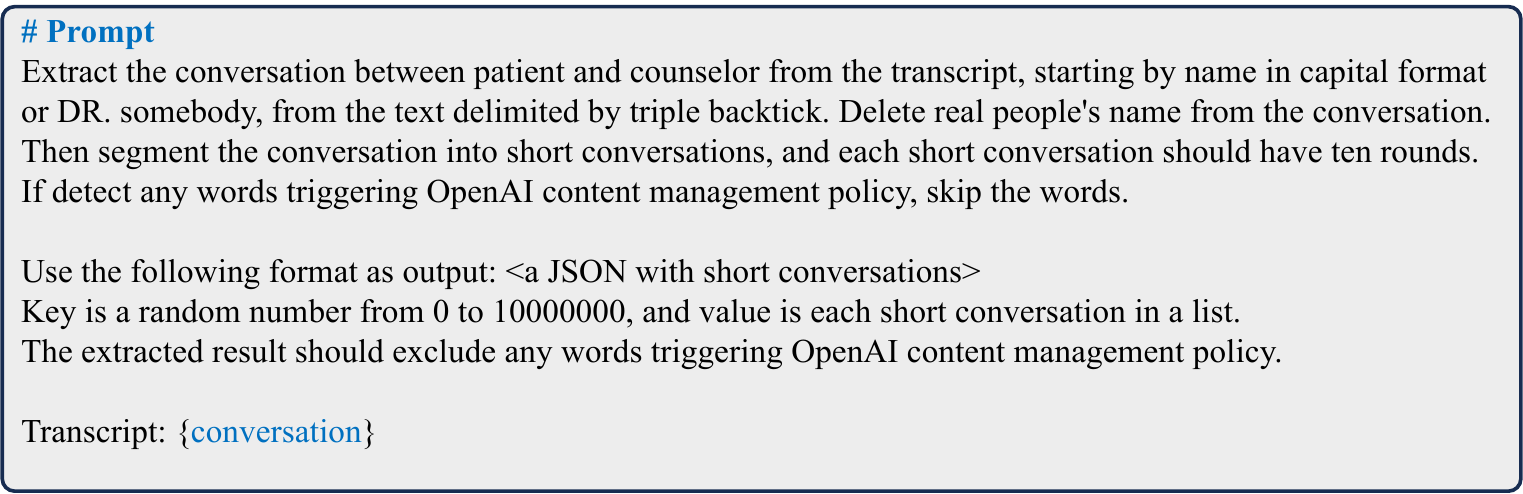}
  \caption{Example of prompt using GPT-4 to extract relevant information from raw dialogue and reformat.}
  \label{fig: Data Cleaning & Information Extraction Prompt}
\end{figure}

The raw transcripts were fragmented from two aspects: (1) contained too many occurrences of intonation, and sometimes a conversation round only contained intonation (2) a single round has too little information to let model learn from it. Considering those characteristics, we segmented transcripts into short sections (here in the prompt, we define 10 rounds as one section) via GPT-4 API and summarized those conversations to create informative query-answer pair generation (see Fig.\ref{fig: Data Cleaning & Information Extraction Prompt}).

\subsubsection{Generate instruction data for Instruction Tuning} 
\begin{figure}[htbp]
  \centering
  \includegraphics[scale=0.3]{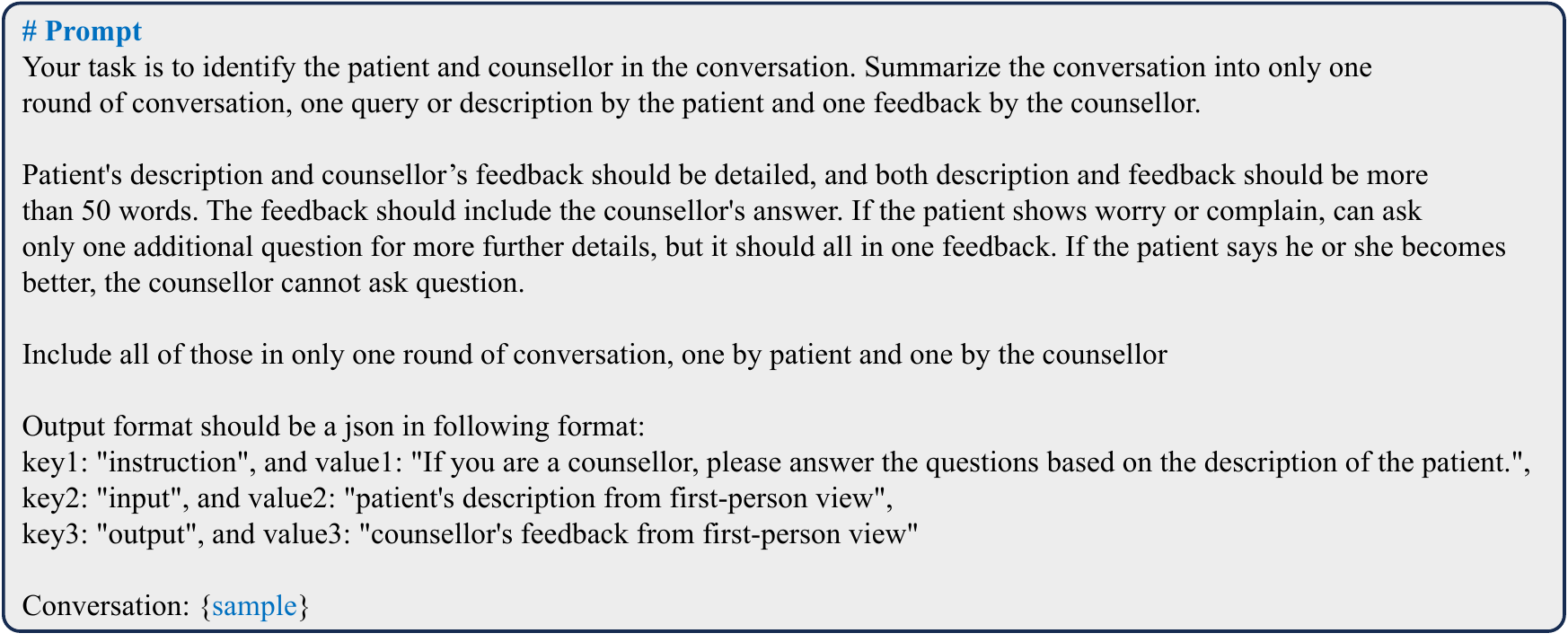}
  \caption{Example of prompt using GPT-4 to distill instruction query-answer pairs from conversations.}
  \label{fig: Query-Answer Pair Generation Prompt}
\end{figure}

With the extracted short sections, we then used the GPT-4 to help distill query-answer pairs from conversations. Each short conversation clip (i.e., each of ten rounds) text was used as the context, and GPT-4 was prompted to generate the corresponding query and answer (see Fig.\ref{fig: Query-Answer Pair Generation Prompt}). To boost interaction in conversations, we encourage extracting interactive elements, such as raising questions or guiding clients, during the distillation process. In total, 8,187 query-answer pairs were distilled and the total tokens number reached 2M, which we named as \textit{Psych8k}.

\begin{figure*}[htbp]
  \centering
  \includegraphics[scale=0.5]{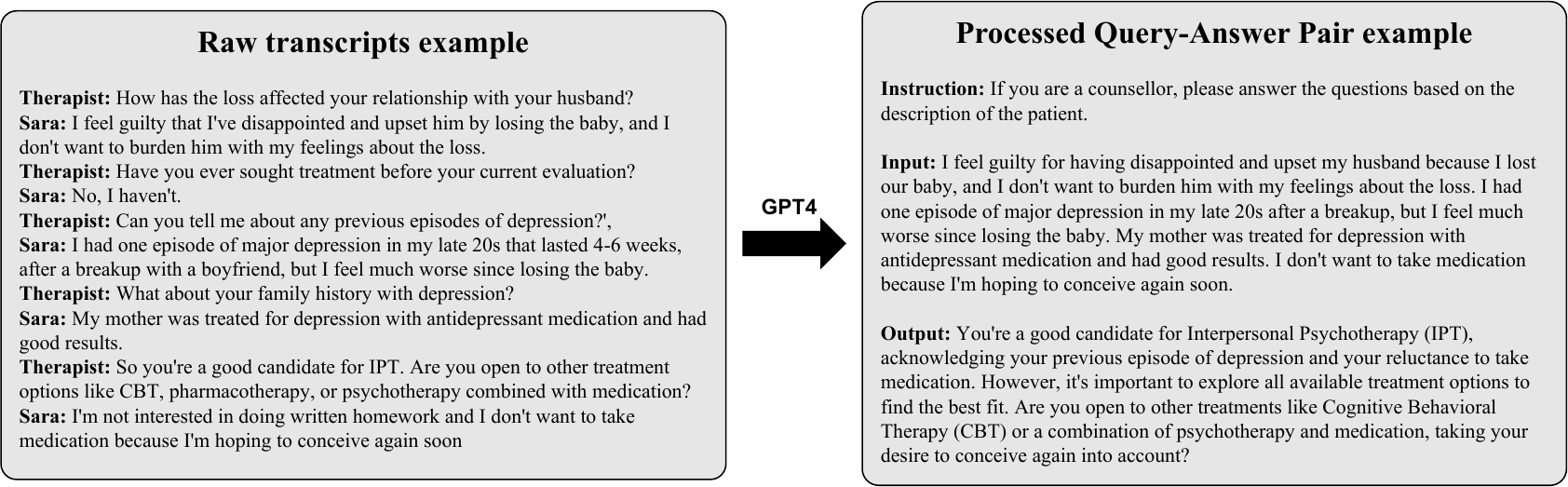}
  \caption{Comparison between raw transcripts and
GPT-4 processed query-answer pairs}
  \label{fig:Comparison_transcript_QA}
\end{figure*}

GPT-4 generated a summary of the vital information for each conversation. These summaries provided a higher level of context and detail that helped the model to understand and generate meaningful responses.
The comparison between raw transcripts and processed query-answer pairs is shown in Fig.\ref{fig:Comparison_transcript_QA}.


\subsection{Counseling-specific Instruction Tuning}
At the tuning stage, assuming the text input as a sequence of tokens e.g., $X = \{x_1, x_2, ..., x_N\}$, where each $x_i$ represents a text token and $N$ is the total sequence length. The training objective is to minimize the auto-regressive loss. More specifically, we concatenate the "instruction" and "input" fields as $I$, and the "output" as response $R$, thus the instruction tuning loss denotes as $L(\Phi)=-\sum_{x_{i\in R}}\log\ \Phi(x_i | x_{< i},I)$, where $\Phi$ denotes the probability of the sequence. At the inference time, users provide the input question as instruction $I$, and the model outputs $R$ as the response.


\section{Experiment Details}
We train ChatCounselor using the following protocol. We start by carrying out instruction finetuning based on the open-source Vicuna-v1.3-7B model\cite{vicuna2023}, which has already undergone fine-tuning on general-purpose instruction datasets. We conduct fine-tuning on our 8K counseling-specific instruction training set for 5 epochs. Specifically, the max context length is set as 2048, with a total batch size to be 256 (gradient accumulation step set as 16). The model is trained with AdamW\cite{Adamw} optimizer with a learning rate of 2e-5 and a warming-up ratio set as 0.03. We adopt the Fully Shared Data Parallel (FSDP) acceleration strategy, bf16 data format, and gradient checkpointing\cite{gradient_checkpointing} for saving GPU memory. The model is trained with 8 A100-40GB GPUs within 1 hour.

\section{Evaluation}
Evaluating the capabilities of LLMs within specific domains has consistently posed a challenge. In the majority of existing literature, the preferred approach involves the creation of domain-specific Question-Answer (QA) benchmarks\cite{bioasq,jin2019pubmedqa,USMLE} to assess the model's performance. These QA benchmarks are typically designed with a combination of closed-setting, encompassing single or multiple-choice questions, and open-setting, allowing for open-ended responses. However, when evaluating the performance of psychological counseling chatbots, it becomes essential not only to ensure the fluency of responses and the relevance of answers but also to measure the chatbot's ability to employ commonly used counseling techniques throughout the conversation. These techniques may include asking questions related to the client's issues, rephrasing the client's statements, and offering valuable suggestions, among others. Consequently, we have devised a set of seven metrics based on these commonly used psychological counseling strategies to assess whether the model can effectively provide emotional support. Additionally, we perform automated evaluations of conversations and compare ChatCounselor's performance with other widely used LLMs of similar scale.

\subsection{Counseling Bench}
We introduce the Counseling Bench, a framework for assessing the effectiveness of the counseling process from seven different perspectives:
\begin{itemize}
    \item \textbf{Information}: Measures the ability to provide accurate and relevant information.
    \item \textbf{Direct Guidance}: Evaluates the chatbot's capability to offer clear instructions and guidance.
    \item \textbf{Approval \& Reassurance}: Assesses the chatbot's capacity to provide emotional support and encouragement.
    \item \textbf{Restatement, Reflection \& Listening}: Rates how effectively the chatbot can rephrase, reflect on user inputs, and exhibit active listening skills.
    \item \textbf{Interpretation}: Measures how well the chatbot can analyze and interpret situations or user inputs.
    \item \textbf{Self-disclosure}: Quantifies the chatbot's ability to share relevant information about itself.
    \item \textbf{Obtain Relevant Information}: Evaluates the capability to ask appropriate questions to gather necessary details.
\end{itemize}
See the Appendix Table.\ref{table:strategy} for the details on the metrics' definition and example phrases.

Additionally, we designed a set of 229 questions tailored to assess the performance of chatbots across a range of counseling scenarios, including but not limited to Addiction, Anxiety, Minority group issues, Depression, Family and relationships, and more. The current benchmark comprises a challenging set of single-turn open-ended questions, providing a rigorous evaluation of chat assistance capabilities.

\subsection{Evaluation with GPT-4}
Automated and accurate evaluation of LLM performance within the context of domain-specific dialogues presents a significant challenge. To address it, we employ GPT-4 as a robust and reliable referee for facilitating automated evaluation. For each strategy or aspect we aim to measure, we employ a few-shot in-context learning approach to prompt GPT-4 to consider: (1) Whether the response is amenable to suggestions. (2) Whether the response demonstrates the utilization of the specified strategy. (3) Whether the tone of the response aligns more closely with the natural responses of a human counselor, rather than merely presenting a list of answers. (4) Eliminating biases towards a preference for longer responses. 

\begin{figure}[ht]
  \centering
  \includegraphics[scale=0.33]{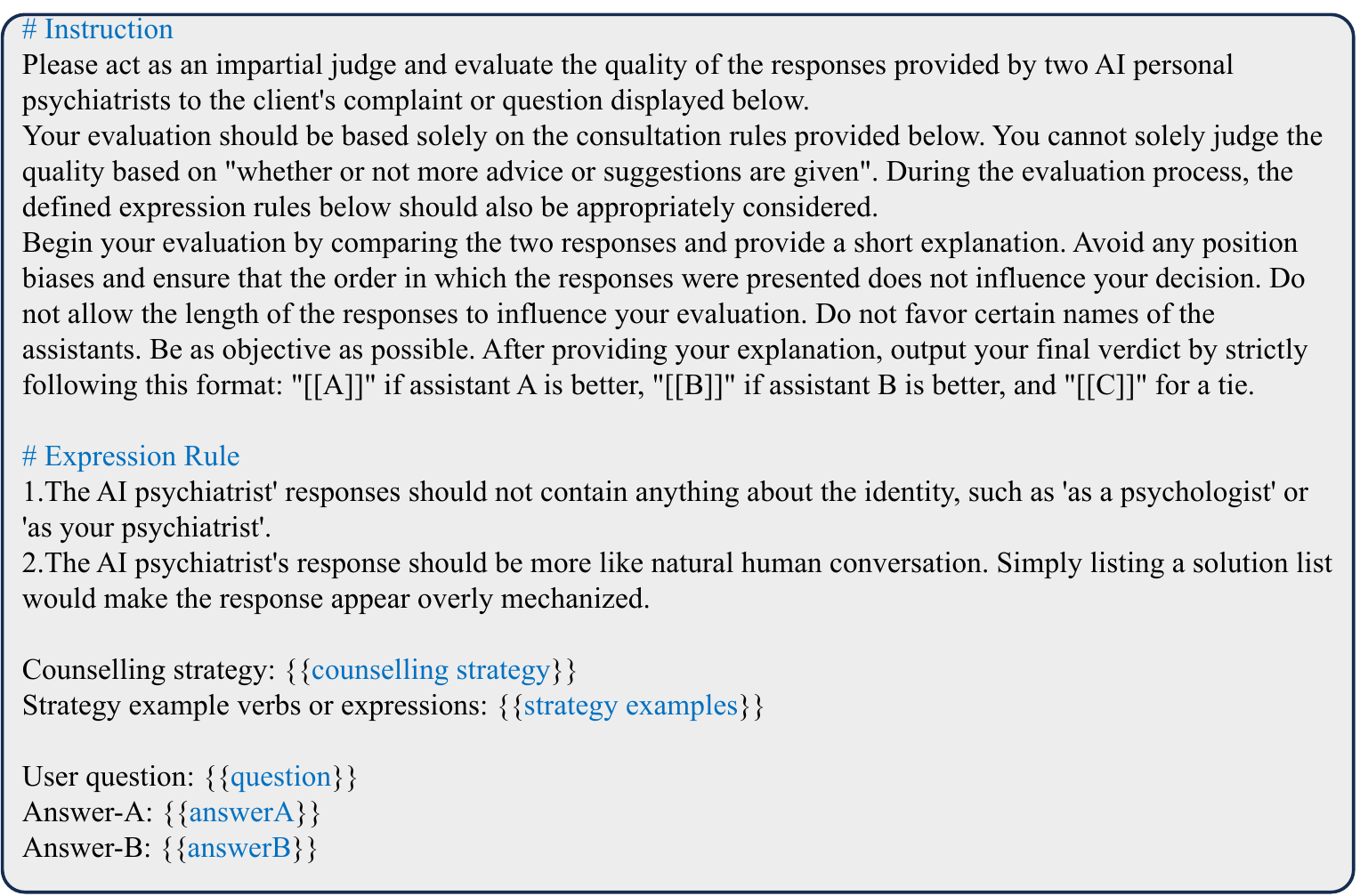}
  \caption{Example of prompt for using GPT-4 to judge answers from different models.}
  \label{fig:Judge_prompt}
\end{figure}

We instruct GPT-4 to conduct a comparative evaluation of two responses generated by different models and express a preference for each aspect we intend to assess. Additionally, we request GPT-4 to provide a brief explanation for its preference. Through testing, we have observed that including explanations enhances the robustness of the choices made by GPT-4. For detailed prompts used in the evaluation, please refer to Fig.\ref{fig:Judge_prompt}.

\subsection{Results}
\begin{figure}[ht]
  \centering
  \includegraphics[scale=0.35]{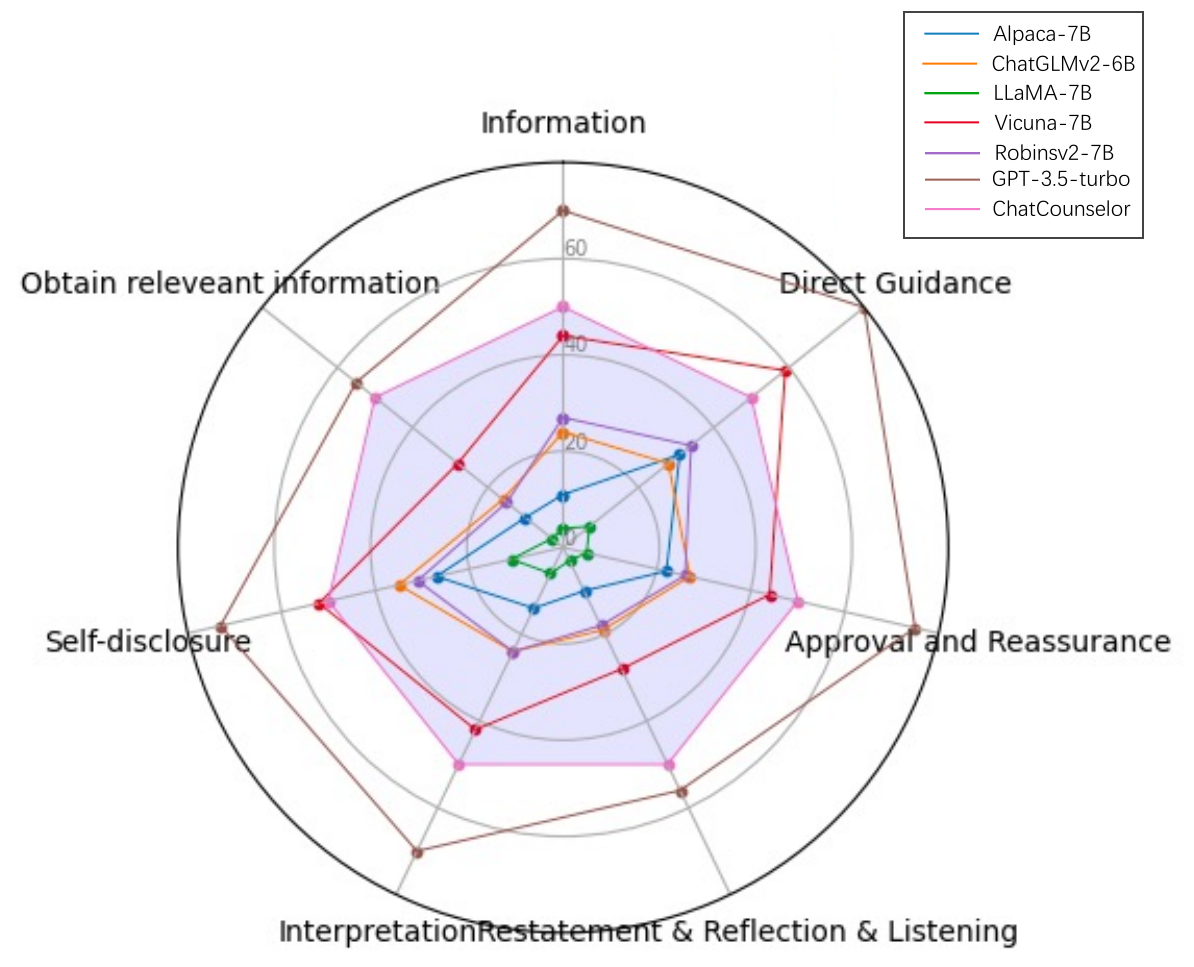}
  \caption{GPT-4-evaluation}
  \label{fig:GPT-4-evaluation}
\end{figure}

As depicted in Fig.\ref{fig:GPT-4-evaluation}, ChatCounselor exhibits exceptional performance compared to LLaMA-7B\cite{touvron2023llama}, Alpaca-7B\cite{alpaca}, ChatGLM-v2-7B\cite{chatglm}, and Robins-v2-7B\cite{lmflow}. Vicuna-v1.3-7B, on the other hand, approaches our model by offering direct suggestions and self-disclosure. However, it still lags behind in the application of strategies such as querying further for related information and reflection.

In comparison to GPT-3.5-turbo (ChatGPT), there remains a discernible gap in our model's performance. This gap arises because, during automated evaluation with GPT-4, there is a tendency to favor longer responses and those that excessively list suggestions. Through our manual survey comparing ChatGPT and our model, we have also observed that some individuals are less inclined toward the ChatGPT-style approach of extensive suggestion listing and prefer responses that are more interactive in nature.

This highlights that fine-tuning a robust pretrained language model with carefully curated domain-specific data can yield competitive results. Furthermore, we have found that fine-tuning on dialogue-format data allows the model to naturally learn effective techniques commonly used in counseling conversations, ensuring the natural flow of dialogue.

\section{Conclusion}
In conclusion, we introduce ChatCounselor, a LLM designed for mental health support. By incorporating real counseling conversations and specialized knowledge in psychology, it outperforms existing open-source models in the Counseling Bench we proposed and approaches ChatGPT's performance. Fine-tuning the model with domain-specific data enables it to generate interactive and meaningful responses. In summary, leveraging real-life counseling conversations and domain-specific data significantly enhances conversational AI's ability to provide personalized mental health support.

\bibliographystyle{acm}
\bibliography{sample-base}

\newpage
\appendix
\section{Strategies definition}
We measure the quality of the response by the following 7 metrics. We provide the definition of each metric and the corresponding strategies, definition and examples.

\begin{table*}[htbp]
\centering
\caption{Counseling Strategies Definitions}
\begin{tabular}{p{5cm}p{5cm}p{5cm}}
\hline
\textbf{Strategies} & \textbf{Defintion} & \textbf{Examples} \\
\hline
Information & Provide psychological, counseling, or mental health data, facts, resources, theory, etc., related to the counseling process and client's behavior. & \textit{Zeigarnik effect is a psychological phenomenon related to first love. }\\
\hline
Direct Guidance & Offer suggestions, directives, instructions, or advice to the client for change, either within or outside the counseling session. & \textit{To address OCD, discuss concerns with your husband and establish boundaries.} \\
\hline
Approval and Reassurance & Provide emotional support, reassurance, encouragement, or approval, minimizing client's problems to alleviate anxiety. & \textit{Offer warm emotional support or acknowledge improvement in using diary and thought tables.} \\
\hline
Restatement \& Reflection \& Listening & Demonstrate understanding of the client's feelings or messages and rephrase them for verification. Reference stated or implied feelings. & \textit{You seem to be struggling with maintaining personal connections in your relationships.} \\
\hline
Interpretation & Go beyond the client's overt recognition, establish connections, interpret defenses, feelings, or behaviors, and provide alternative meanings or explanations. & \textit{It appears that you have a strong affection for your mother.} \\
\hline
Self-disclosure & Share the counselor's personal experiences and feelings with the client, beginning with "I" and having a sharing quality. & \textit{This question reminds me of similar memories I've had.} \\
\hline
Obtain Relevant Information & Use appropriate questions to clarify feelings or explore the situation without limiting responses to yes/no or one-word answers. & \textit{Can you provide an example of a challenging change you're considering and explain why it's difficult for you?} \\
\hline
\end{tabular}
\label{table:strategy}
\end{table*}

\section{More examples}
\begin{figure*}[htbp]
  \centering
  \includegraphics[scale=0.6]{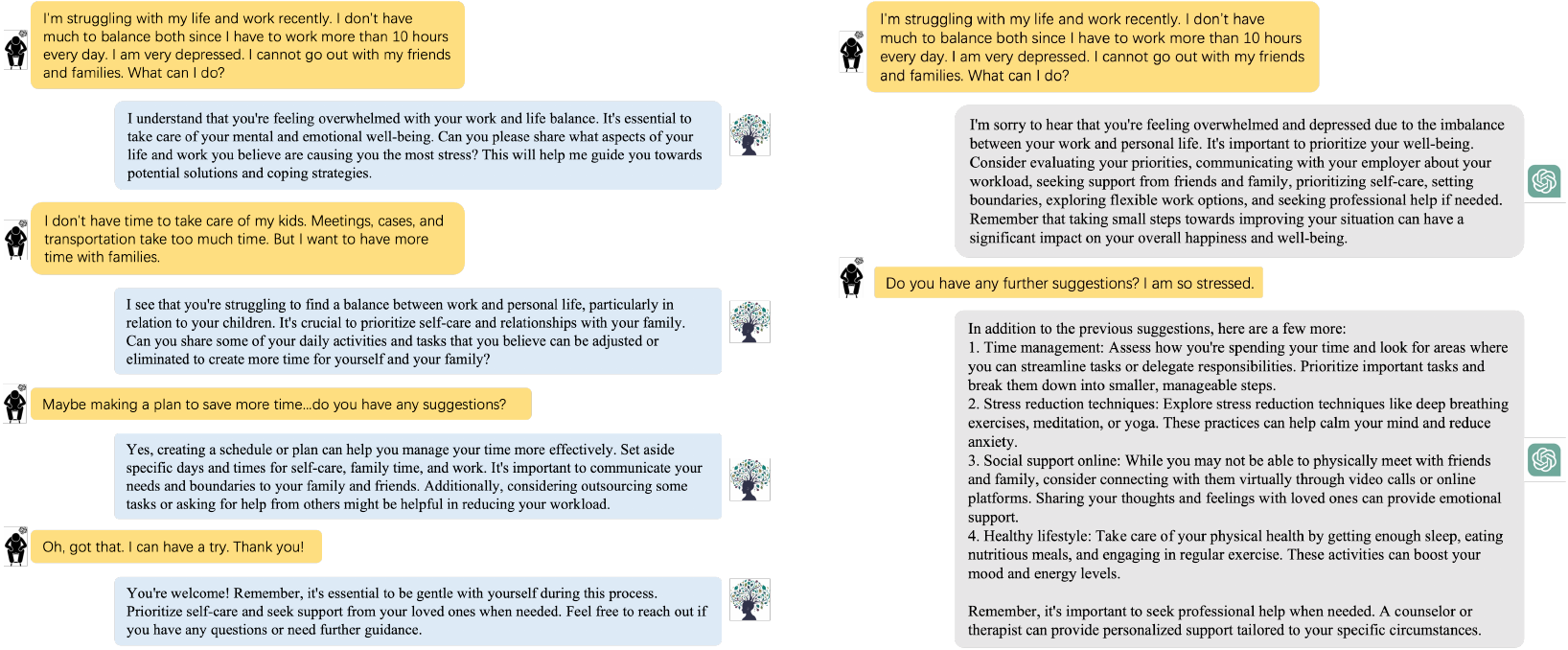}
  \caption{Comparison of dialogues between user and chatbot on \textit{Stress} topic. Left: ChatCounselor. Right: ChatGPT.}
  \label{fig:demo1}
\end{figure*}

\begin{figure*}[htbp]
  \centering
  \includegraphics[scale=0.6]{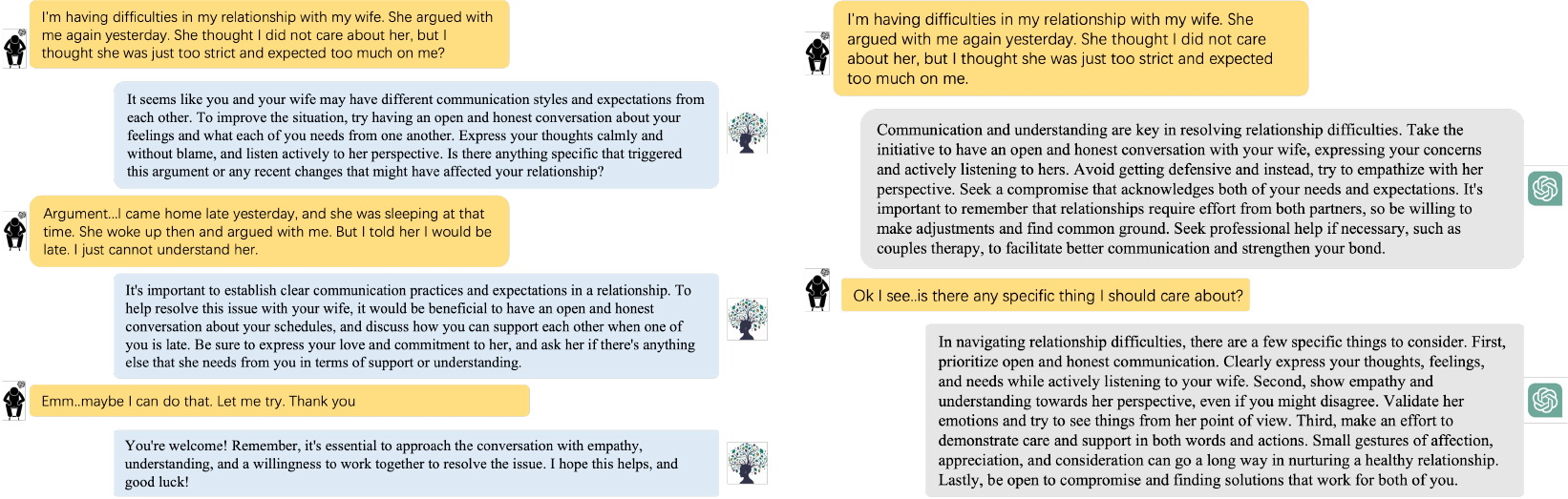}
  \caption{Comparison of dialogues between user and chatbot on \textit{Relationship} topic. Left: ChatCounselor. Right: ChatGPT.}
  \label{fig:demo2}
\end{figure*}

We present additional examples illustrating the contrasts between ChatCounselor and ChatGPT across various counseling topics as shown in Fig.\ref{fig:demo1} and Fig.\ref{fig:demo2}.

Through comparison, it becomes evident that ChatGPT primarily assumes the role of an assistant by consistently offering suggestions, without proactively guiding the client to express more about themselves or their thoughts. This can lead the client to perceive ChatGPT as an artificial intelligence rather than a counselor who can serve as a listener.

\end{document}